\documentclass[10pt,a4paper,conference]{IEEEtran}
\usepackage{cite}
\usepackage{listings}
\usepackage[utf8]{inputenc}

\usepackage{float}
\usepackage{amsmath,amssymb,amsfonts}
\usepackage{algorithmic}
\usepackage{graphicx}
\usepackage{textcomp}
\def\BibTeX{{\rm B\kern-.05em{\sc i\kern-.025em b}\kern-.08em
    T\kern-.1667em\lower.7ex\hbox{E}\kern-.125emX}}

\usepackage{color}
\definecolor{codegreen}{rgb}{0,0.6,0}
\definecolor{codegray}{rgb}{0.5,0.5,0.5}
\definecolor{codepurple}{rgb}{0.58,0,0.82}
\definecolor{backcolour}{rgb}{0.95,0.95,0.92}
\DeclareMathAlphabet{\pazocal}{OMS}{zplm}{m}{n}

\newcommand{\listingsfont}{\ttfamily}

\lstdefinestyle{mycodestyle}{
    backgroundcolor=\color{backcolour},
    commentstyle=\color{codegreen},
    keywordstyle=\color{magenta},
    numberstyle=\scriptsize\color{codegray},
    stringstyle=\color{codepurple},
    basicstyle=\footnotesize\linespread{0.9}\listingsfont,
    breakatwhitespace=false,
    breaklines=true,
    captionpos=b,
    keepspaces=true,
    numbers=left,
    numbersep=5pt,
    showspaces=false,
    showstringspaces=false,
    showtabs=false,
    tabsize=2
}
\begin{document}

\title{An Overview of Robotic Grippers}

\author{}

\maketitle

\section*{Authors}
Mr Thomas J. Cairnes is a PhD candidate at University of Bristol. His main area of research is the grasping and perception of soft, deformable objects with underactuated hands. (pd18523@bristol.ac.uk)

Mr Christopher J. Ford is a PhD candidate at the University of Bristol. His research interests are focused on the use of tactile perception and feedback for real-time grasping and manipulation of multi-fingered robotic hands. (chris.ford@bristol.ac.uk)

Dr Efi Psomopoulou is an Assistant Professor in the Dept. of Engineering Mathematics at the University of Bristol. Her research interests are physical robot interaction, grasping and manipulation. (efi.psomopoulou@bristol.ac.uk)

Prof. Nathan Lepora is a Professor of Robotics and AI in the Dept. of Engineering Mathematics at the University of Bristol. His research interests are robot dexterity, biomimetics and tactile sensing (n.lepora@bristol.ac.uk)   

\section{Introduction}
The development of robotic grippers is driven by the need to execute particular manual tasks or meet specific objectives in handling operations. Grippers with specific functions vary from being small, accurate and highly controllable such as the surgical tool effectors of the Da Vinci robot (designed to be used as non-invasive grippers controlled by a human operator during keyhole surgeries) to larger, highly controllable grippers like the Shadow Dexterous Hand (designed to recreate the hand motions of a human). Additionally, there are less finely controllable grippers, such as the iRobot-Harvard-Yale (iHY) Hand or Istituto Italiano di Tecnoglia-Pisa (IIT-Pisa) Softhand, which instead leverage natural motions during grasping via designs inspired by observed bio-mechanical systems.

As robotic systems become more autonomous and widely used, it is becoming increasingly important to consider the design, form and function of robotic grippers.

\section{Basics of Robotic Grippers}
While the designs of robotic grippers are varied, there are common descriptors that can be used to separate them into groups that are informative about their operating principles.

Anthropomimetic grippers aim to replicate the motion of the human hand, and are inspired by the success of human dexterity to provide a familiar reference point for intended motion. Alternatively, biomimetic but non-anthropomimetic grippers take inspiration from other bio-mechanical structures to inform their motion and control. Lastly, there are non-biomimetic grippers. This category has not been investigated as much as grippers inspired by the huge range of diversity in natural organisms but still provides examples of effective grippers.

There are four general types of actuation in modern robotic grippers: pneumatic grippers driven by compressed air, hydraulic grippers driven by compressed fluid, electric grippers using motors and vacuum grippers relying on some form of suction.

As mentioned in the introduction, it is important to consider grippers in two overarching categories: \textit{Fully Actuated} or \textit{Underactuated} grippers. These terms indicate the degree of control that can be expected from a gripper. A fully-actuated gripper has every Degree of Freedom (DOF) directly controlled by a motor or other actuation mechanism. Conversely, an underactuated gripper has only some DOFs actively controlled while the rest react passively according to the mechanical structure of the gripper.

\begin{figure}
  \includegraphics[width=\linewidth]{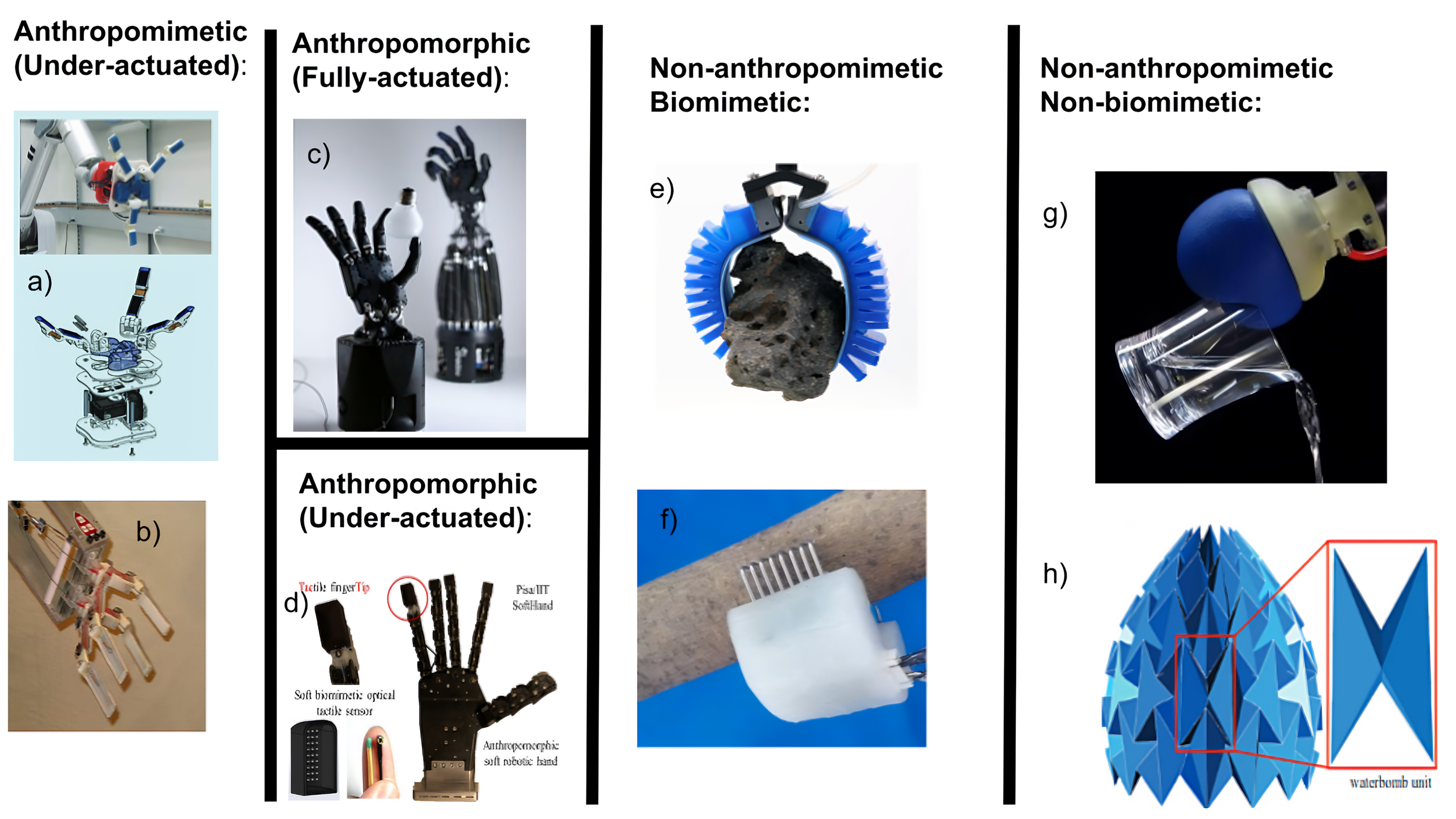}
  \caption{\textbf{Examples of Robotic Grippers.} a) The Yale Openhand Model O: A 3-fingered anthropomimetic hand open-sourced from customising the iHY Hand and developed at Yale University (Dollar \textit{et al} 2017). b) The SDM Hand: A 4-fingered anthropomimetic gripper utilising compliant passive joints, developed at Harvard University (Dollar \textit{et al} 2007). c) The Shadow Dexterous Hand: A fully actuated 5-fingered anthropomorphic hand developed by the Shadow Robotics Company. d) The Pisa/IIT Softhand: A 5-fingered anthropomorphic hand utilising rigid-passive links and a single actuator to achieve grasps; this version has a biomimetic tactile fingertip developed at Bristol Robotic Laboratory (Lepora \textit{ et al.} 2021). e) A gecko-nspired adhesive gripper: A biomimetic gripper developed to utilise electro-static forces to adhere to objects, developed at the University of California (Glick \textit{et al} 2018). f) An inchworm-inspired soft gripper: The claws mimic the muscle structure of an inchworm for climbing robots, developed in Guangdong University China (Li \textit{et al} 2017). g) A Universal Jamming Gripper: A gripper designed to use the unique jamming properties of granular materials to grasp a range of objects, developed at Cornell University (Brown \textit{et al} 2010). h) An Origami-inspired folding gripper: This gripper uses rigid-flexible links to grasp, developed at The Harbin Institute of Technology (Liang \textit{et al} 2021)}
\end{figure}

The simplest robotic grippers are fully actuated two-DOF grippers, often referred to as parallel jaw grippers. The earliest was mounted on the Stanford Arm (1969) and was originally hydraulic but later operated through DC motors. In the following decades, industrialised adaptations of parallel jaw grippers have been used widely in various picking and packing applications and are still widely used today because of how straightforward and capable they are to use. 

\subsection*{Anthropomimetic Grippers}
This section focuses on anthropomimetic grippers, {\em i.e.} grippers that use opposing multi-jointed fingers to form grasps similar to those humans may perform. 

An early example of an anthropomimetic gripper was the Graspar hand (1996), a 3-fingered hand comprised of mostly passive joints driven by an internal pulley system. The Schunk and Barret 3-fingered hands (2007) were rigid and fully actuated. This made them very controllable, however they had issues when applied to a wide range of objects as their motions required precise control of every joint, which made them difficult to control. These hands relied almost exclusively on power grasps; {\em i.e.} grasps where the hand seeks to completely envelope an object, as opposed to a precision grasp where contact with the object is made with the fingertips of the hand.

A desire for smooth anthropomimetic motions to reduce control complexity, led to underactuated hands utilising compliant passive joints. The 4-fingered Shape Deposition Manufacturing (SDM) Hand (2010) was constructed almost entirely of soft compliant materials all driven by a single motor. Its design focused on creating anthropomimetic power grasps that maximise contact between the fingers and an object, with the finger configuration resembling two soft, compliant parallel jaw grippers. Passive, dislocatable joints enabled hand designs that could produce natural and stable power grasps. In this way, one joint may stop moving after making contact with an object, and the connected joints in the chain can continue to move with the actuation until they contact the object. This produces a grasp that maximises the contact area without prior knowledge of the object, which is a property commonly referred to as being \textit{adaptive}. 

There then followed hands optimised for power grasps using passive joints such as Robotiq's Adaptive Gripper (2013) and Kinova's Jaco (2013). However, those adaptive grippers were still fairly rigid in construction and had not capitalised on the strengths shown by the SDM hand. 

Consequently, the iHY Hand (2013) was designed for a range of specific tasks to be achieved, not just grasping objects. This additional functionality required the ability to perform precision and power grasps, requiring individually actuated fingers and enough rigidity to perform tasks like flicking switches. The iHY utilised 5 motors but was underactuated as it still had 8 DOFs, making use of a rigid-flexible construction for compliant fingers. The iHY hand was well suited for research applications, being simply fabricated, and was open sourced as the GRABLAB Model-O while additionally being commercialized as the Reflex Hand from RightHand Robotics.

Full anthropomorphism is achieved when a hand is 5-fingered, for example, the Pisa/IIT SoftHand (2014) is a semi-soft, underactuated gripper intended to be easily controllable whilst also applicable to a variety of grasping tasks. A single motor results in the hand only having one grasping mode (a whole hand enveloping grasp) with tendons routed through the hand, resulting in a grasp that follows a ‘postural synergy’ which closely mimics the motion of a human enveloping grasp. The rigid-flexible construction, dislocatable joints and anthropomimetic design result in a gripper that is highly adaptive yet simple to control. More recent versions have included additional control from a second postural synergy, and an easily-fabricated 3D-printed version of the SoftHand has been released, known as the BRL/Pisa/IIT SoftHand (2022).

At the opposite extreme in terms of actuation is the Shadow Dexterous Hand, a highly actuated anthropomorphic gripper demonstrated as capable of replicating human grasps during teleoperation due to its highly actuated nature. This has been demonstrated in a teleoperation task where an operator solved a Rubik's Cube using a pair of the hands.

Anthropomimetic grippers offer a familiarity of control and design not found in other types. Most notably due to the human hands development focused on tool use and task completion. While the capabilities of other biomimetic grippers are much more specialised, anthropomimetic grippers offer a much wider range of application cases.

\subsection*{Biomimetic Non-anthropomimetic Grippers}
Traditionally, robotic grippers have been comprised of rigid links and hard components, which offer precise control, better overall actuation strength and efficient transfer of power through the system. However, natural systems make use of rigid-flexible couplings, for example in the joints of the human body. One of the earliest examples of a soft robotic gripper utilising rigid-flexible links is Hirose's soft gripper (1978), which resembled an Octopus tentacle, and was the first gripper to passively adapt itself to an object's shape. Such rigid-flexible links would later become an important component in some anthropomimetic grippers such as the Pisa/IIT SoftHand.

There are also many examples of entirely compliant grasping solutions in nature, such as suction cups on cephalopod tentacles or adhesive ant-eaters tongues. Non-anthropomimetic grippers inspired by mechanisms like these have long been widely studied in the application of Soft Robotics, leading to grippers that depart from the soft-rigid principle. For example, NASA's gecko gripper (2016) is based around gecko feet for holding flat objects in space. The microscopic hairs on a gecko's feet generate intermolecular forces (van Der Waals forces), allowing adhesion without any liquid or surface tension. NASA's gripper operates on the same principle, making it suitable for moving and orientating large panels or objects that would otherwise require an large gripper to power grasp that would otherwise not be well-suited for space applications.

As discussed earlier these biomimetic grippers offer grasping capabilities that are highly specialised. Capable of grasping and manipulating objects that anthropomimetic grippers couldn't. This, however, comes with its own issues regarding design, control and application of these grippers.
\subsection*{Non-biomimetic Grippers}
There are currently only a few examples of grippers that are not based on the human hand or other biomechanical structures in living organisms. For example, vacuum grippers used suction cups to create vacuums to adhere to an object, which are reminiscent of cephalopod sucker's but were not inspired by them. Early vacuum grippers used industrial vacuums and cycling airflow to provide constant suction, but progressed quickly to more energy-efficient equivalents of vacuum gripping technology. In particular, rather than cycling air for constant suction, the vacuum gripper presses against an objects surfaces then has the air in the cup expelled, creating a pressure imbalance for adhesion. 

An interesting example of a non-biomimetic gripper is the universal 'jamming' universal gripper utilising the phenomena for granular materials to 'jam', {\em i.e.} pack together very tightly when forced together. In one design an elastic container (like a balloon) is filled with a granular material, pushed against an object then vacuum suction removes the air from the balloon and jams the material together, which maintains the same shape but is much stiffer so it can grasp the object securely.

The end-effectors of surgical robots such as the Da Vinci or Smart Tissue Autonomous Robot (STAR) are miniaturised surgical tools that can have gripping functions. For some applications, like surgical pliers, they can be considered grippers, but in others they are more specialised tools like scalpels.

Grippers in this category are once again specialised even further for specific applications but lend themselves to being applied and controlled easily. Vacuum grippers and jamming grippers are incredibly simple and straightforward but cannot accomplish the more complex tasks that anthropomimetic grippers are aiming towards.
\section{Grasp Perception}
An important capability for robotic grippers is to perceive they interacting with objects as they come into contact or are holding the object. This could be comprised of interoceptive sensing inside the hand itself due to pressure, friction, tension in actuated components, which are analogous to the human perception of grasping through prioprioceptive and other signals. This information combined with exteroceptive information such as from vision creates a feedback loop for interacting with objects and forms the building blocks of an automated system for a robotic gripper to interact with the world.

This separation of perception into visual and tactile components is important because grasping does not entirely concern sensing in the gripper alone: knowledge of the gripper exclusively is not enough to make intelligent decisions regarding interacting with objects or completing tasks.
\subsection*{Proprioception}
Proprioception encompasses a suite of senses internal to the body that enables humans to accurately estimate the position of their own limbs, head and trunk without being able to see them. These senses originate from muscles, tendons and joint ligaments to detect changes in skeletal muscle contraction/stretch, body/limb position and joint capsule stress.

The proprioceptive sense of a robotic gripper depends on the type of gripper. For example, a two-finger gripper driven by electric motors can sense proprioceptively by using encoders in the motor, with measurements of how far a servo has rotated and how much torque is applied being common indications of effective grasping. A gripper driven pneumatically or hydraulically, however, can present challenges for integration with proprioceptive sensing. For example, the previously mentioned Graspar hand had no proprioceptive capability at all. Simply measuring that the actuator or piston is active provides more detail, but additional sensors such as barometers are usually needed for detailed proprioceptive information. 

Fully-actuated robotic grippers such as the Shadow Dexterous Hand have controllable servos for each joint in the fingers. Hence, from the encoder readings on the servos, an operator or autonomous system can ascertain the state and position of each finger. 
\subsection*{Exteroception}
Exteroception is the sensing of external stimuli and can be seen as a system-level capability of the entire robot rather than for the gripper alone. By utilising cameras and depth sensors along with a model of the platform, the system can localise itself and construct a representation of its environment. %Combining this with proprioceptive information will synthesise a complete perception of the working environment and the system itself.
\subsection*{Tactile Sensing}
The human sense of touch is essential for our grasping and dexterous capabilities, relating specifically to sensations carried through the skin. Tactile sensing allows humans to evaluatea broad range of physical contact properties including shape, texture, softness, temperature and noxious stimuli (pain). It is also used in combination with proprioception to make estimates of weight and how much force is required to grasp an object. One can view touch as forming the sensory boundary between proprioception and exteroception. Therefore, when progressing towards intelligent grippers, the role of tactile sensing and the link it provides between outward and inward perception is completely crucial.

Tactile sensors range in size and complexity but their baseline function is the same; determining whether the gripper has made contact with something. From simple pressure sensors that enable contact checking to piezoelectric tactile arrays, to optical tactile sensors such as the TacTip or GelSight sensors. Optical tactile sensors rely on a camera observing what is usually a representation of a layer of human skin. The TacTip traces white Papillae on the inside that deflect and displace in response to surface stimuli. The GelSight traces shading and dots printed directly under the `skin' to observe stretch and topography changes at a small scale.

\section{Levels of Autonomy}
The research field of Autonomous Robotics considers Levels of Autonomy (LoA) that separate into 6 groups, which delineate where the control and responsibility of an operator lie in relation to the type of autonomous system.

\subsection*{Human Control \& Teleoperated - LoA 0-1}
In any area where a gripper must interact with a human being, it is generally assumed that most interactions are under human control, whcih can be referred to as `Kineasthetic Guidance'. This may either be through direct control of an operator or through the system being prompted by a human to act then wait for further instructions. There must also be sufficient feedback to the human operator to allow reliable control.

A shared autonomy situation refers to tasks where a robot may operate autonomously but final control of the system rests with a human. For example, a prosthetic robot hand could be LoA 1 if the hand is capable of regulating itself to close around an object but requires human instruction and decision making to do so. %In this situation the hand regulating itself to close and remain closed around an object is a simple automated function putting it into LoA 1.

The Da Vinci surgical robot is directly controlled by a human operator, who may be in the same room, via teleoperation. Such surgical robots have closed-loop control systems that offer stable control of surgical tools. This capability differentiates them from being under direct kineasthetic guidance. Likewise, solutions for hazardous/dangerous environments are often teleoperated in a similar manner.

\subsection*{Shared Autonomy - LoA 2-3}

A progression in the level of autonomy would be, for example, a warehouse scenario where a human inputs an item request for the robot to collect from some storage bin. The arm manoeuvring to the bin, identifying the item, determining how to grasp it and retrieving it would be automated. However a human still controls when the task is undertaken, the details of the task, and can make a decision to stop or cancel the task. Such operations are at LoA 2.

Giving such systems the capability to initiate tasks autonomously in a self-determined order to just retrieve and deliver items, would give the robot sufficient autonomy to reach LoA 3. The human operation is limited to interventions during emergencies and ceasing operations. In such scenarios a robot can handle some safety features itself, but is still ultimately dependent on a human operator. Many driverless cars can be viewed as LoA 3.

\subsection*{Fully Automated Situations - LoA 4-5}
This full LoA is where there is no human involvement at all. Achieving this LoA does not imply that the system is safe, it just means that the system is responsible for its own safety measures.

Consider, for example, the automated warehouse described above and imagine that a section of the warehouse has been closed off for the robot to operate independently, containing storage bins and containers for certain item types that the system is capable of handling autonomously. The robot has lists of tasks that it performs without supervision, making all relevant decisions regarding the collection of items and the item order. This is a fully autonomous system that has been limited to operate in a specific scenario that is LoA 4. This scenario separation may be because of the operational limits of the system itself; for example, the grippers on the arms may not be able to handle objects larger than the grippers, so the system is constrained to prevent it from attempting tasks on which it may fail.

If instead of limiting the item types or section of the warehouse the system can operate in, it is allowed to operate for all items types in the entire warehouse then a LoA 5 system has been created. Fully automated shipping yards are good examples of LoA 5 systems, as they can handle all standard shipping containers and do so without supervision from a human.
\begin{figure}
  \includegraphics[width=\linewidth]{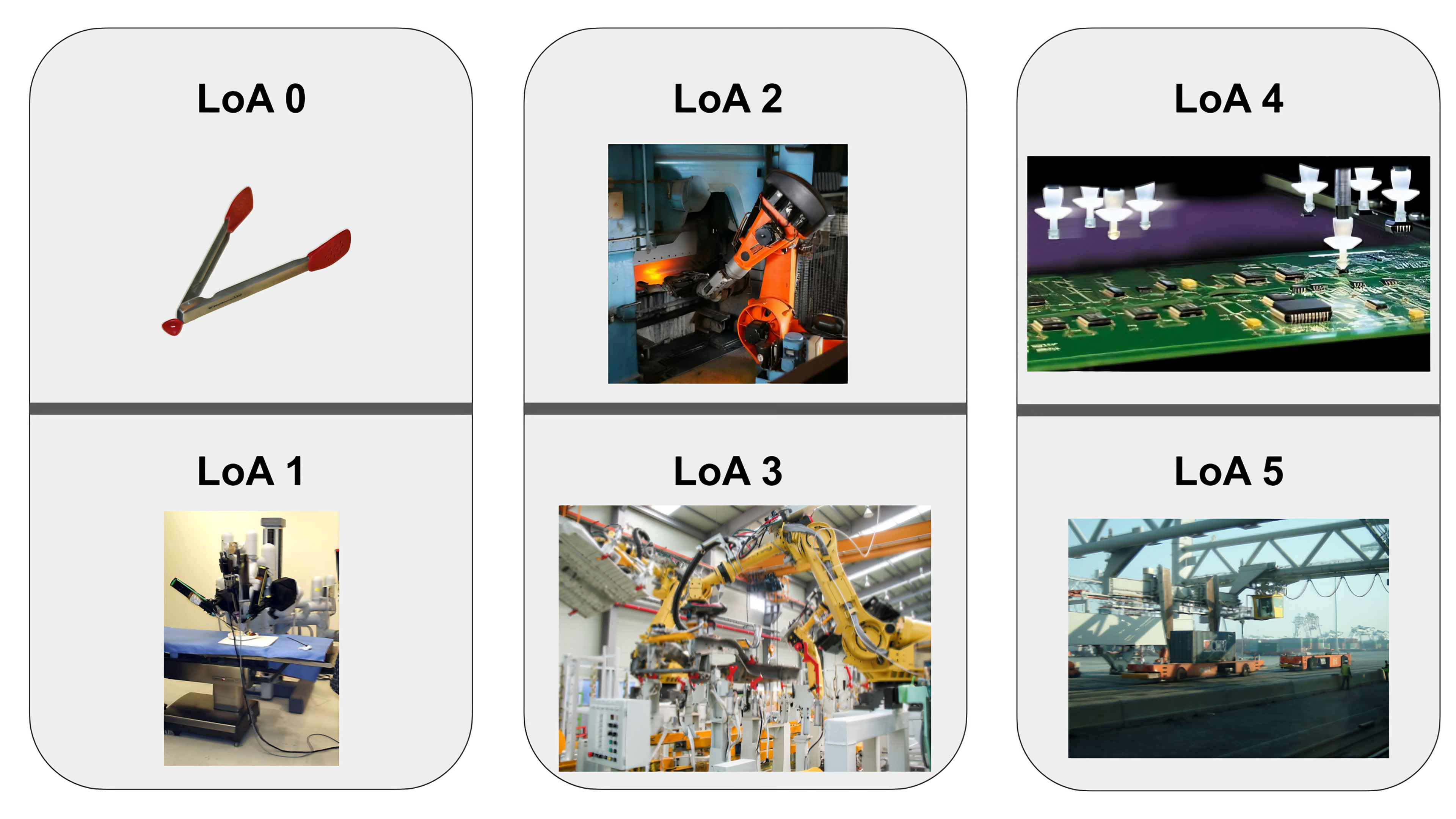}
  \caption{\textbf{Examples of Levels of Autonomy.} LoA 0: A pair of Kitchen tongs, a gripper with no input or control on the gripper side. LoA 1: A Da Vinci Surgical robot, teleoperated with feedback control loops to eliminate instrument shake. LoA 2: An autonomous robot used in a smelting factory, waits for human input then performs a pre-programmed set of instructions LoA 3: A factory robot assembling cars. Beginning of operation and emergency stops are still under human control. LoA 4: A Surface Mounted Device pick and place system using  a pneumatic gripper to assemble devices. The machine operates in a workspace separate from humans and thus has control over its own safety measures. LoA 5: The Port of Rotterdam is an almost entirely autonomous shipping port, automated stacking cranes perform necessary grasping. In this situation the system can accept tasks, fulfil them, access resources it needs autonomously and can reasonably grasp any object it will encounter..}
\end{figure}
\section{Ethical Considerations}
\subsection*{Safety and Harm}
The ethical considerations of a robotic gripper being used on or nearby humans is important but is complicated by the gripper being just one component of an entire robotic  system. In many regards, safety standards that apply to systems like robotic arms can also be applied to robotic grippers.

In the early development of workplace and factory robots, the simplest way to ensure human safety was to have a clearance zone that a person should not enter, or to isolate the robot entirely within a cage. However, separating humans and robotics is proving to be limiting in modern applications of robotics.

The development of collaborative industrial robots has brought them into human workspaces, demanding a much higher degree of safety to allow close proximity to humans. Safety measures can involve: monitored safety stops, {\em i.e.} a stopping operation when a human enters the workspace; speed and separation monitoring that causes the robot to slow operations when humans are nearby; and power limiting. Power and force limiting requires either monitoring of actuation mechanisms or limiting through inherent design. For example, soft grippers in collaborative scenarios offer a straightforward and implicit method of providing operational safety limits. .

Since the primary function of a gripper is to secure the position and orientation of an object in relation to the robotic platform, a reasonable safety standard would be to require that a gripper should not change state upon losing power. A gripper dropping something fragile, heavy or sharp during an emergency stop or power cut could be harmful and/or costly.

\subsection*{Impacts on Employment}
Developing grippers that are more capable of performing like human hands, or perhaps are superior to humans for specific situations/tasks, will lead to development of autonomous platforms that can perform tedious/dangerous tasks in place of humans. How this affects the job market is a matter of debate, as the shifting economic growth and recession of the past decade has made it difficult to evaluate the impact that automating labour has actually had. An important question is whether increasing automatisation of human physical labour would simply remove jobs or create more skilled and fulfilling work in other areas such as the installation and maintenance of those robots?

Whatever the answer, future robotic technologies should be used responsibly and to improve human society. One clear responsible use in this scenario would be to target jobs and tasks that are considered dangerous or tedious, so that the humans doing those jobs can be safer and focus on more productive or creative work.

\section{Future Challenges}
\subsection*{Moving further towards Intelligent Grasping}
A key problem for future of robotic grasp is how to grasp intelligently. Modern robotic hands have become highly dexterous and articulated but the challenge of controlling them remains unsolved.

The use of more sophisticated and human-like sensing in a robotic gripper will lend itself to enabling intelligence, but also creates a need to process and interpret more data on top of a potentially already very complex gripping system. For example, during the development of the iHY hand in the early 2010s, highly articulated dexterous hands were available and had the same issues that they still face now in regards to uptake and application in industrial spaces. Simpler grippers are currently a more accessible option, but have their own issues with integration of greater sensing capacity into structurally simple hands, often require a complete change in design to be compatible with a given sensing technology; for example, vision-based tactile sensors are difficult to integrate into soft grippers.

\subsection*{Universality vs Specificity}
So-called universal grippers offer a cheap and effective way of grasping many objects, containing a particle material enclosed in fine rubbery membrane for enclosing around objects.However, they can fail to offer a reliable solutions for many tasks, such as grasping in clutter or object manipulation. Likewise, a gripper designed for a specific task, such as 2-fingered grippers for picking and placing simple objects, may fail to work outside of that task. So far, the workaround has been to change the tasks you give to certain grippers or limit their application space. While this separation has enabled progress, it has led to a huge diversity in gripper types, The development of a gripper that can be used universally and controlled accurately, which to some extent the human hand solves, is still ongoing.
\subsection*{Unstructured Environments}
Unknown, unstructured and cluttered environments are problematic for any robotic system. Finding ways to explore spaces and dealing with clutter is not only a challenging perception task but also requires intelligent planning. The system needs to be capable of differentiating between unimportant clutter that can be ignored or displaced, and dangerous disturbances that must be safely dealt with. High proprioceptive and tactile sensing capabilities are needed for gripping systems that can operate in such environments.

In terms of levels of autonomy as we discussed earlier, at present LoA 5 systems exist in very predictable and structured environments such as shipping yards.
\subsection*{Safety}
While collaborative robots are beginning to make more of an industrial appearance their uptake remains slow. From a legal and practical standpoint, it is  currently far simpler and safer for company to isolate their robots from human workers. It remains an important challenge to improve the productivity of humans working alongside robots while maintaining their safety. As grippers develop towards more intelligent grasping alongside human works, their operational safety will remain a key consideration.
\section{Conclusion}
The development of robotic grippers has advanced enormously over the past few decades, which is exhibited by the diverse range of manipulator designs and applications. The next steps for the use of robotic grippers needs to address the fact that highly articulated hands are still very costly, difficult to control and have not seen as much application as their potential suggests. Currently, cheaper and less complex grippers are finding more industrial uses. The developers of dexterous robot hands need to make their grippers easier for commercial users to utilise, otherwise those users will stick with current options and use-case scenarios. Moreover, further advancements in the perceptual and control capabilities of simpler grippers are much-needed stepping stones to integrating intelligent gripping technologies into fully autonomous systems.

%Greater uptake of dexterous hands is needed before chall%enges regarding their use and high entry cost can be solved. Until then
%\section*{Read More}
 \begin{itemize}
 \item Young, B. "The Bionic-Hand Arms Race: High-Tech Hands are Complicated, Costly, and Often Impractical." \textit{IEEE Spectrum} 59.10: 24-30, 2022.
 \item Odhner L., Jentoft L., Claffee M., Corson N., Tenzer Y., Ma R., Buehler M., Kohout R., Howe R., Dollar a. A compliant, underactuated hand for robust manipulation. \textit{The International Journal of Robotics Research}, 33(5):736–752, 2014.
 \item Dollar A. and Howe R. The highly adaptive SDM hand: Design and performance evaluation. \textit{International Journal of Robotics Research}, 29(5):585–597, 2010.
 \item Ward-Cherrier, B., Pestell N., Cramphorn L., Winstone B., Giannaccini M., Rossiter J., and Lepora N. "The TactTip family: Soft optical tactile sensors with 3d-printed biomimetic morphologies." Soft robotics 5, no. 2: 216-227, 2018.
 \item Ma, R. and Dollar A. "Yale OpenHand Project: Optimizing Open-Source Hand Designs for Ease of Fabrication and Adoption" \textit{IEEE Robotics `I\&' Automation Magazine (Volume: 24, Issue: 1, March 2017) Pages: 32-40}
 \item Dollar A. and Howe R. "The SDM Hand as a Prosthetic Terminal Device: A Feasibility Study". \textit{2007 IEEE 10th International Conference on Rehabilitation Robotics}, 13-15 June 2007.
 \item Lepora N., Ford C., Stinchcombe A., Brown A., Lloyd J., Catalano M., Bianchi M., Ward-Cherrier B. "Towards integrated tactile sensorimotor control in anthropomorphic soft robotic hands". /textit{2021 IEEE International Conference on Robotics and Automation (ICRA)}. 30 May 2021 - 05 June 2021 
 \item Glick P., Suresh S., Ruffatto D., Cutkosky M., Tolley M., Parness A. "A Soft Robotic Gripper With Gecko-Inspired Adhesive". IEEE Robotics and Automation Letters ( Volume: 3, Issue: 2, April 2018) Pages:903-910. 
 \item Li M., Su M., Xie R., Zhang Y., Zhu H., Zhang T., Guan Y. "Development of a bio-inspired soft gripper with claws". 2017 IEEE International Conference on Robotics and Biomimetics (ROBIO), 05-08 December 2017. 
 \item Brown E., Rodenberg N., Amend J., Mozeika A., Steltz E., Zakin M., Lipson H., Jaeger H. "Universal robotic gripper based on the jamming of granular material". Proceedings of the National Academy of Sciences 107(44) 2010
 \item Liang D., Gao Y., Huang H., Li B. "Design of a Rigid-Flexible Coupling Origami Gripper". 2021 IEEE International Conference on Robotics and Biomimetics (ROBIO), 27-31 December 2021.
 \end{itemize}
%\nocite{*}

\end{document}